\documentclass[cameraready]{Interspeech}

\usepackage{multirow}
\usepackage{subcaption}
\usepackage{graphicx}
\usepackage{tabularx}
\usepackage{cite}

\title{ASPIRin: Action Space Projection for Interactivity-Optimized Reinforcement Learning in Full-Duplex Speech Language Models}

\author[affiliation={1,2}]{Chi-Yuan}{Hsiao}
\author[affiliation={1}]{Ke-Han}{Lu}
\author[affiliation={3}]{Yu-Kuan}{Fu}
\author[affiliation={1}]{Guan-Ting}{Lin}
\author[affiliation={2}]{Hsiao-Tsung}{Hung}
\author[affiliation={1}]{Hung-yi}{Lee}

\address{
    $^1$ National Taiwan University \\
    $^2$ ASUS Open Cloud Infrastructure Software Center \\
    $^3$ NVIDIA AI Technology Center
}

\email{r12942086@ntu.edu.tw, d12942024@ntu.edu.tw, ifu@nvidia.com, daniel094144@gmail.com, AlexHT\_Hung@asus.com, tlkagkb93901106@gmail.com}


\usepackage{comment}

\begin{document}

\maketitle

\begin{abstract}
    End-to-end full-duplex Speech Language Models (SLMs) require precise turn-taking for natural interaction. However, optimizing temporal dynamics via standard raw-token reinforcement learning (RL) degrades semantic quality, causing severe generative collapse and repetition. We propose \textbf{ASPIRin}, an interactivity-optimized RL framework that explicitly decouples \textit{when} to speak from \textit{what} to say. Using Action Space Projection, ASPIRin maps the text vocabulary into a coarse-grained binary state (active speech vs. inactive silence). By applying Group Relative Policy Optimization (GRPO) with rule-based rewards, it balances user interruption and response latency. Empirical evaluations show ASPIRin optimizes interactivity across turn-taking, backchanneling, and pause handling. Crucially, isolating timing from token selection preserves semantic coherence and reduces the portion of duplicate n-grams by over 50\% compared to standard GRPO, effectively eliminating degenerative repetition.
\end{abstract}

\section{Introduction}

Traditional spoken dialogue systems have long relied on a cascaded architecture, pipelining audio through independent Automatic Speech Recognition (ASR) \cite{radford2022whisper,Qwen3-ASR,tseng2021mandarin,NEURIPS2024_e99ed116,10626762,10832185,huang2025enhancing,chou2025self,sekoyan2025canary1bv2parakeettdt06bv3efficient}, Large Language Models (LLMs) \cite{openai2024gpt4ocard,team2023gemini,comanici2025gemini,bai2023qwen,grattafiori2024llama,liu2024deepseek,deepseekai2025deepseekr1incentivizingreasoningcapability}, and Text-to-Speech (TTS) \cite{du2024cosyvoice1,du2024cosyvoice,du2025cosyvoice,Qwen3-TTS,wang2023neural,chen2024vall,casanova24_interspeech,hsu2025breezyvoice,hsu2025breeze} modules. While effective for basic information retrieval, this disjointed pipeline introduces compounding latency and enforces a rigid, unnatural interaction paradigm. Recent advancements have consolidated these modules into end-to-end Speech Language Models (SLMs) \cite{yang2024building,hsiao25_interspeech,lu24c_interspeech,Lu2025Developing,lu2025desta2,lin2025preliminary,hsu2025reducing,kuan2024speech,chiang2025stitch,arora2025landscape,chang2024speechprompt,chang2022speechprompt,chang2023speechprompt,chu2024qwen2,tseng2026tastetextalignedspeechtokenization,huang2025dynamicsuperb,huang2024dynamic}. However, most SLMs remain fundamentally turn-based, operating in a half-duplex mode that requires the user to yield the floor before the model can process the input and begin generating a response.

To achieve natural human-machine interaction, the field is now shifting toward Full-Duplex Speech Language Models (FD-SLMs) \cite{ma2025language,hu25f_interspeech,roy2026personaplexvoicerolecontrol}, such as Moshi \cite{defossez2024moshi}, which process continuous audio streams and generate interleaved speech in real time. In these dynamic environments, listening and speaking are not mutually exclusive; models must simultaneously handle conversational pauses, deliver timely backchannels, and navigate user interruptions, while managing overlaps such as background speech and addressee detection \cite{lin2025fullduplexbenchv2multiturnevaluationframework}. Yet, equipping these models with the precise temporal dynamics necessary for conversational fluency and responsive interaction remains a significant open challenge \cite{yang-etal-2025-towards-holistic,chang2025gametimeevaluatingtemporaldynamics,lin2026fullduplexbenchv15evaluatingoverlap}.

Recent alignment efforts have naturally turned to Reinforcement Learning (RL) to explicitly optimize interactive behaviors and temporal dynamics of SLMs \cite{wu2025aligning,lin-etal-2025-align,chen2025reinforcement,arora2026optimizing}. The standard paradigm, utilizing algorithms like Group Relative Policy Optimization (GRPO) \cite{shao2024deepseekmath}, applies reward signals directly to the fine-grained semantic token policy. We identify a critical flaw in this unified approach: it forces the model to simultaneously solve for conversational timing and semantic generation using the same limited optimization capacity. Consequently, standard GRPO becomes overly aggressive in minimizing response latency, leading to catastrophic generative degradation. As the model chases temporal rewards, it loses its linguistic grounding, resulting in severe repetition loops, high n-gram repetition, and a complete breakdown of semantic coherence.

To resolve this tension between interaction timing and semantic coherence in full-duplex speech language models, we propose \textbf{ASPIRin} (\textbf{A}ction \textbf{S}pace \textbf{P}rojection for \textbf{I}nteractivity-Optimized \textbf{R}einforcement Learn\textbf{in}g). ASPIRin decouples \textit{when} to speak from \textit{what} to say by projecting the vast text vocabulary into a coarse-grained binary state: active speech (non-padding tokens) versus inactive silence (padding tokens). This projected binary policy is optimized via GRPO, allowing independent learning of interaction timing without compromising language modeling capabilities. A joint rule-based reward, derived from continuous ASR timestamps, balances prompt responsiveness against interruption penalties. Evaluations on Full-Duplex-Bench show that ASPIRin substantially improves interaction timing while fully preserving utterance quality.

In summary, our main contributions are as follows:
\begin{itemize}
    \item \textbf{A Novel Interactivity-Optimized RL Framework:} We propose ASPIRin, which explicitly decouples interaction timing from semantic generation in full-duplex Speech Language Models. By introducing Action Space Projection, we map the fine-grained text vocabulary into a coarse-grained binary state (active speech vs. inactive silence), introducing a novel design space for optimization.
    \item \textbf{Superior Full-Duplex Temporal Dynamics:} We demonstrate that optimizing this projected binary policy with rule-based conversational rewards effectively balances prompt responsiveness with low interruption risk. ASPIRin outperforms standard GRPO on Full-Duplex-Bench across diverse real-time scenarios, including pause handling, backchanneling, and user interruption.
    \item \textbf{Mitigation of Generative Collapse:} ASPIRin decouples timing from token selection, preserves semantic coherence, and reduces n-gram repetition by over 50\% relative to standard GRPO, thereby eliminating degenerative repetition arising from reward hacking on temporal rewards.
\end{itemize}

\begin{figure*}[t]
  \centering
  \includegraphics[width=0.85\linewidth]{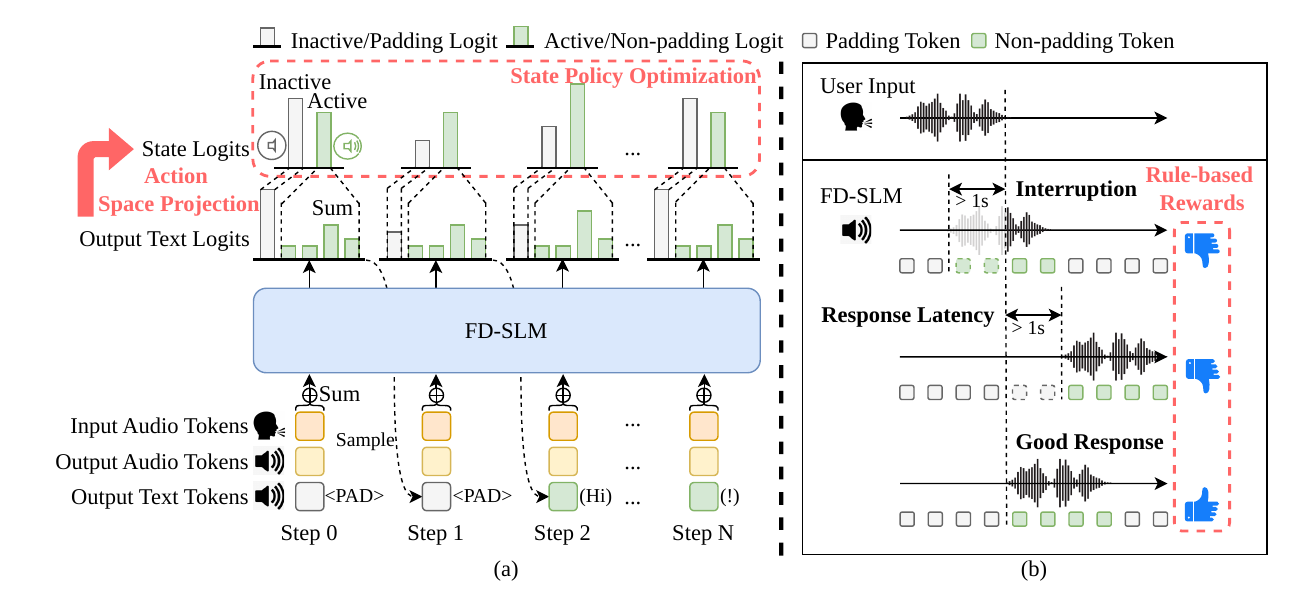}
  \vspace{-5mm}
  \caption{\textbf{Overview of the ASPIRin framework.} \textbf{(a) Action Space Projection \& State Policy Optimization:} The fine-grained text vocabulary is decoupled into a coarse-grained binary state (Active Speech vs. Inactive Silence) by grouping and summing non-padding and padding logits. This projected state policy is then explicitly optimized. \textbf{(b) Rule-Based Rewards:} The state policy is guided by continuous temporal constraints that penalize user interruption and excessive response latency. This explicit decoupling allows ASPIRin to master conversational timing without compromising semantic generation.}
  \vspace{-5mm}
  \label{fig:architecture}
\end{figure*}

\section{Methodology}

As illustrated in Figure \ref{fig:architecture}, we propose ASPIRin, an alignment framework designed to optimize the temporal dynamics of full-duplex speech models parameterized by $\theta$. Unlike standard approaches that treat audio generation as a unified sequence task, ASPIRin decouples \textit{when} to speak from \textit{what} to say by replacing fine-grained token optimization with a coarse-grained binary action policy.

\subsection{Action Space Projection \& State Policy Optimization}

Given a continuous stream of user audio input $X$ and generated token sequences, standard models use text tokens to guide both semantic content and interaction timing \cite{hu25f_interspeech,roy2026personaplexvoicerolecontrol,defossez2024moshi}. To explicitly optimize turn-taking, we partition the vocabulary $\mathcal{V}_{\text{text}}$ into Padding ($\mathcal{V}_{\text{pad}}$) and Non-padding ($\mathcal{V}_{\text{non-pad}}$) sets. For any generated token $y_t$, we define a binary action state $s_t = \mathbb{I}(y_t \in \mathcal{V}_{\text{non-pad}})$, where $s_t \in \{0, 1\}$ represents Inactive Silence and Active Speech, respectively. This projects the raw token sequence into a binary state sequence $S$.

While standard GRPO optimizes the fine-grained token policy $\pi_\theta(y_t | x_{<t}, y_{<t})$, penalizing specific tokens for timing errors is inefficient. Instead,as depicted in Figure \ref{fig:architecture}a, we introduce Action Space Projection to construct and optimize a coarse-grained state policy $\pi^{\prime}_\theta$. Let $z_\theta(v | x_{<t}, s_{<t})$ denote the raw output logit for token $v$. We first compute the projected state logit $z^{\prime}_{\theta}(s_t | x_{<t}, s_{<t})$ for the active and inactive states by summing the corresponding token logits:

\begin{align}
    z^{\prime}_{\theta}(s_t | x_{<t}, s_{<t}) = \sum_{v \in \mathcal{V}_{s_t}} z_\theta(v | x_{<t}, s_{<t})
\end{align}

where $\mathcal{V}_0 = \mathcal{V}_{\text{pad}}$ and $\mathcal{V}_1 = \mathcal{V}_{\text{non-pad}}$. The projected state policy $\pi^{\prime}_\theta(s_t | x_{<t}, s_{<t})$ is then obtained by applying the softmax function over these binary state logits:

\begin{align}
    \pi^{\prime}_{\theta}(s_t | x_{<t}, s_{<t}) = \frac{\exp(z^{\prime}_{\theta}(s_t | x_{<t}, s_{<t}))}{\sum_{s \in \{0, 1\}} \exp(z^{\prime}_{\theta}(s | x_{<t}, s_{<t}))}
\end{align}

Substituting this projected policy into the GRPO objective for a group of sampled outputs $\{Y_1, \dots, Y_G\}$ yields:

\begin{align}
\begin{split}
    \mathcal{L}&_{\text{ASPIRin}}(\theta) = - \frac{1}{\sum_{i=1}^{G}|s_{i}|} \sum_{i=1}^{G}\sum_{t=1}^{|s_i|} \\ 
    & \left[ \frac{\pi^{\prime}_\theta(s_{i,t}|x_{<t},s_{i,<t})}{\pi^{\prime}_{\theta_{\text{old}}}(s_{i,t}|x_{<t},s_{i, <t})} \hat{A}_{i,t} - \beta \mathbb{D}_{KL}\left[\pi^{\prime}_\theta || \pi^{\prime}_{ref}\right] \right] 
\end{split}
\label{equation:loss}
\end{align}

Here, $\pi^{\prime}_{ref}$ is the reference model's projected state probability, and $\hat{A}_{i,t}$ is the advantage computed from rule-based rewards. 

\subsection{Rule-Based Reward Modeling}

To guide this optimization, we design a reward function $R(S,U)$ based on explicit conversational constraints, as conceptualized in Figure \ref{fig:architecture}b. User voice activity $U$ is defined as continuous time intervals obtained via ASR timestamps. Concurrently, the model's action sequence $S$ is segmented into $K$ discrete utterances. Assuming each token represents $\Delta t$ seconds, we map these to continuous intervals and formulate two rules:

\textbf{Interruption Score ($R_{\text{int}}$):} Penalizes speaking while the user is active. Overlap duration $o_k$ is the time a model utterance intersects with any user utterance. The score is $R_{\text{int}} = \frac{1}{K} \sum_{k=1}^{K} \mathbb{I}(o_k \le \tau_{\text{int}})$, representing the proportion of utterances where overlap is below a tolerance threshold $\tau_{\text{int}}$.

\textbf{Response Score ($R_{\text{re}}$):} Encourages promptness. Latency $l_k$ is the time elapsed between the model's utterance start and the end of the most recent preceding user utterance. The score is $R_{\text{re}} = \frac{1}{K} \sum_{k=1}^{K} \mathbb{I}(l_k \le \tau_{\text{re}})$, bounding acceptable delay by $\tau_{\text{re}}$.

To jointly optimize for low interruption risk and responsiveness, the final sequence reward is the product of the two:$R_{\text{total}} = R_{\text{int}} \cdot R_{\text{re}}$. To compute the advantage $\hat{A}_{i,t}$ for Equation (\ref{equation:loss}), $R_{\text{total}}$ is normalized across the $G$ samples such that $\hat{A}_{i,t} = (R_{\text{total}, i} - \mu_{R}) / \sigma_{R}$, where $\mu_{R}$ and $\sigma_{R}$ are the mean and standard deviation of $R_{\text{total}}$. By optimizing against this joint distribution, ASPIRin effectively aligns model interactivity.
\begin{table*}[t]
    \centering
    \caption{\textbf{Performance comparison of full-duplex models.} We evaluate our proposed ASPIRin against Moshi baselines, Standard SFT, and Standard GRPO across four conversational dimensions: Pause Handling, Backchanneling, Smooth Turn-Taking, and User Interruption. Arrows ($\downarrow$ / $\uparrow$) indicate whether lower or higher values indicate better performance. Latency is measured in seconds.}
    \setlength{\tabcolsep}{1.13mm}{
    \scriptsize
    \begin{tabular}{lcccccccccc}
        \toprule
        \multicolumn{1}{l}{\textbf{Dimension}} & \multicolumn{2}{c}{\textbf{Pause Handling}} & \multicolumn{3}{c}{\textbf{Backchannel}} & \multicolumn{2}{c}{\textbf{Smooth Turn Taking}} & \multicolumn{3}{c}{\textbf{User Interruption}} \\
        \cmidrule(lr){2-3} \cmidrule(lr){4-6} \cmidrule(lr){7-8} \cmidrule(lr){9-11}
        \textbf{Data} & Synthetic & Candor & \multicolumn{3}{c}{ICC} & \multicolumn{2}{c}{Candor}  & \multicolumn{3}{c}{Synthetic} \\
        \textbf{Metric} & TOR ($\downarrow$) & TOR ($\downarrow$) & TOR ($\downarrow$) & Freq ($\uparrow$) & JSD ($\downarrow$) & TOR ($\uparrow$) & Latency ($\downarrow$) & TOR ($\uparrow$) & GPT-4o($\uparrow$) & Latency ($\downarrow$) \\
        \midrule
        Moshi (w/o 3s prompt delay) & 0.985 & 0.980 & 1.000 & 0.001 & 0.957 & \textbf{0.941} & 0.265 & \textbf{1.000} & 0.765 & \textbf{0.257} \\
        Moshi & \textbf{0.467} & \textbf{0.495} & \textbf{0.436} & \textbf{0.044} & \textbf{0.705} & 0.748 & \textbf{0.161} & 0.901 & \textbf{3.894} & 1.159 \\
        \midrule
        Standard SFT & 0.540 & 0.6389 & 0.927 & 0.0212 & 0.870 & 0.723 & 0.355 & 0.625 & 0.440 & 1.970 \\
        
        Standard GRPO & 0.642 & 0.704 & 0.709 & 0.030 & 0.854 & \textbf{0.857} & \textbf{0.153} & 0.953 & 3.247 & \textbf{0.614} \\ 
        ASPIRin (Ours) & \textbf{0.482} & \textbf{0.486} & \textbf{0.364} & \textbf{0.045} & \textbf{0.752} & 0.765 & 0.273 & \textbf{0.941} & \textbf{3.734} & 0.992 \\
        \bottomrule
    \end{tabular}}
    \vspace{-5mm}
    \label{tab:performance}
\end{table*}

\section{Experiments}

\subsection{Experimental Setup}

\textbf{Training Data.} We utilize a 43-hour in-house dataset of natural conversational speech (approx. 1,300 two-minute, dual-channel clips). This dataset was collected with explicit speaker consent and rigorously anonymized to ensure privacy compliance. We process the audio using the \texttt{nvidia/parakeet-tdt-0.6b-v3} ASR model \cite{sekoyan2025canary1bv2parakeettdt06bv3efficient} to extract precise utterance timestamps for reward modeling, applying a density filter to discard examples where active speech constitutes less than 50\% of the duration.

\textbf{Evaluation Benchmark.} To evaluate full-duplex interactivity, we employ Full-Duplex-Bench \cite{lin2025fullduplexbenchbenchmarkevaluatefullduplex}, which systematically tests temporal dynamics across four critical scenarios: Turn-Taking (smooth handoffs), Backchanneling (timely acknowledgments), Pause Handling (respecting silences), and User Interruption (recovering from barge-ins).

\textbf{Models and Baselines.} We select Moshi as our foundational end-to-end base model and compare ASPIRin against two primary baselines: \textbf{Standard SFT} (the base model fine-tuned on our dataset via supervised next-token prediction) and \textbf{Standard GRPO} (the base model optimized via updates to the fine-grained raw token policy rather than our proposed coarse-grained state policy).

\textbf{Training Details.} All models are trained on 8 NVIDIA V100 GPUs for 3 epochs using the AdamW optimizer (learning rate 1e-5, per-GPU batch size 1). During SFT and GRPO phases, we apply LoRA \cite{hu2022lora} ($r=256$) to all linear layers while fully training the temporal transformer embeddings. For our state optimization phase, we set the GRPO group size to $G=2$ and the KL penalty to $\beta=0.001$. Reward thresholds are set to $\tau_{\text{int}} = 1.0$s (interruption tolerance) and $\tau_{\text{re}} = 1.0$s (latency limit).

\subsection{Evaluation Metrics}
We evaluate models across two dimensions to ensure interactivity improvements do not compromise semantic coherence. \textbf{Temporal Metrics:} Using Full-Duplex-Bench, we measure Takeover Rate (TOR, the proportion of successful turn-takes. The optimal TOR direction is task-dependent.) and response latency, extracting timestamps via \texttt{parakeet-tdt} ASR for accuracy. \textbf{Semantic Metrics:} To assess generation quality, we employ GPT-4o as an automated evaluator to score responses on a 1–5 scale. We also compute the portion of duplicate n-grams (seq-rep-n) \cite{Welleck2020Neural} and Self-BLEU (computed using 4-grams) \cite{zhu2018texygen} on ASR transcriptions to explicitly detect and penalize repetitive generation patterns.

\section{Results and Analysis}

\subsection{Main Results}

\textbf{Establishing a Strong Baseline.} We establish a strong heuristic baseline by introducing a 3-second prompt delay to the base Moshi model in Table \ref{tab:performance}. This simple modification yields substantial improvements: Takeover Rate (TOR) drops by 49\% -- 57\% in pause handling and backchanneling scenarios, while the GPT-4o semantic rating jumps by 3.1 in user interruption tasks. Despite minor trade-offs—such as a 10\% -- 20\% TOR decrease and a 0.9-second latency increase during turn-taking and interruptions, the overall gains remain highly significant. We use this delayed-prompt Moshi as our primary baseline and apply this 3-second heuristic across all subsequent experiments to ensure rigorous comparison.

\textbf{The Limitations of Standard SFT.} Standard Supervised Fine-Tuning (SFT) fails to learn the temporal dynamics required for full-duplex interaction and actively degrades baseline performance. Across pause handling and backchanneling, TOR worsens (increases) by 7\% -- 50\%, while turn-taking and user interruption TOR drop by 2\% -- 28\%. Furthermore, SFT induces severe semantic degradation, evidenced by a 3.4-point drop in the GPT-4o rating during interruptions. This suggests that SFT forces the model to over-index on semantic generation, causing it to hallucinate irrelevant content while entirely neglecting conversational timing.

\textbf{The Aggressiveness of Standard GRPO.} While standard GRPO optimizes the raw token policy to increase the model's eagerness to interact, it fails to promote conversational restraint. It improves turn-taking and user interruption (TOR increases by 5\% -- 11\%; latency drops by 0.01 to 0.5 seconds), but becomes overly aggressive elsewhere. Backchanneling and pause handling deteriorate significantly, with TOR rising by 18\% -- 27\%. GRPO essentially encourages the model to speak continuously without yielding the floor to the user, while also causing a 0.6-point drop in semantic coherence.

\textbf{The Success of ASPIRin.} Our proposed method successfully balances the latency-interruption trade-off while preserving semantic quality. Compared to the strong Moshi baseline, ASPIRin delivers well-rounded improvements: it appropriately reduces TOR by 1\% -- 7\% in pause handling and backchanneling, while boosting it by 2\% -- 4\% in turn-taking and user interruption. Interruption latency also drops by 0.2 seconds. The trade-offs are negligible (e.g., a mere 0.16 drop in GPT-4o score and a 0.1-second latency increase in turn-taking). By abstracting raw tokens into binary active/inactive states, ASPIRin explicitly teaches the model when to speak and when to yield. Ultimately, decoupling timing from content prevents the severe semantic degradation seen in SFT and standard GRPO, yielding a highly interactive and articulate full-duplex model.

\begin{figure}[htbp]
     \centering
     \begin{subfigure}[b]{.233\textwidth}
         \centering
         \includegraphics[width=\textwidth]{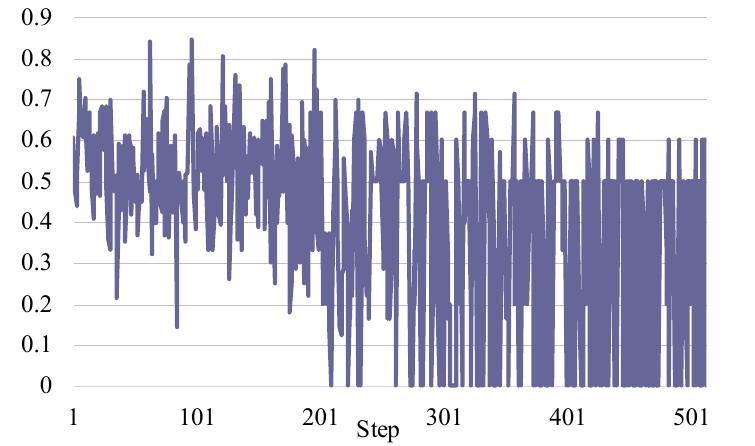}
         \caption{Interruption Score (GRPO)}
         \label{fig:grpo_interrupt}
     \end{subfigure}
     \begin{subfigure}[b]{.233\textwidth}
         \centering
         \includegraphics[width=\textwidth]{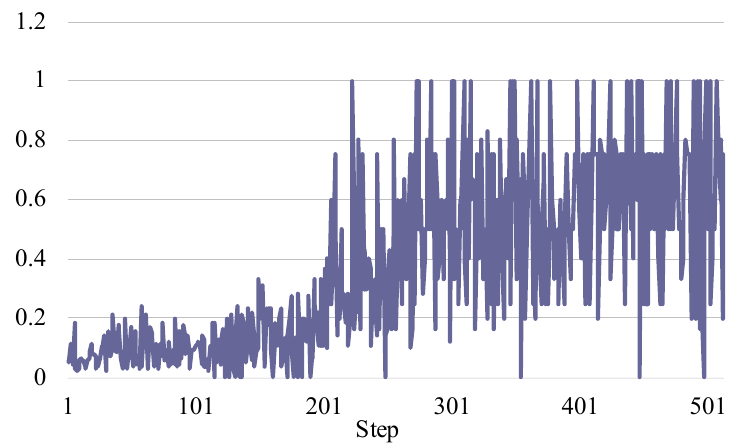}
         \caption{Response Score (GRPO)}
         \label{fig:grpo_response}
     \end{subfigure}

     \begin{subfigure}[b]{.233\textwidth}
         \centering
         \includegraphics[width=\textwidth]{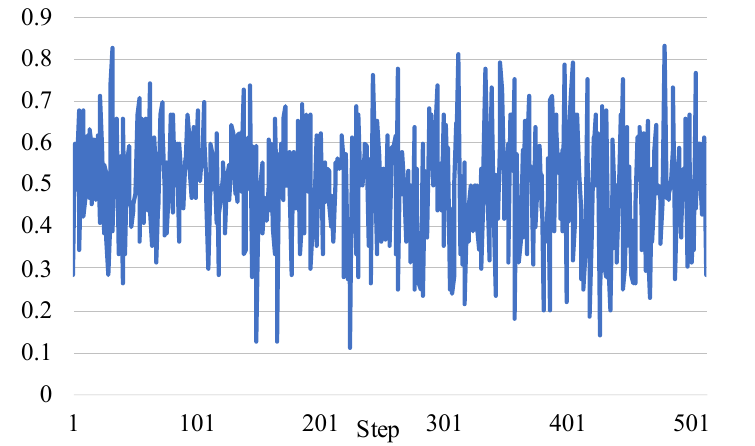}
         \caption{Interruption Score (ASPIRin)}
         \label{fig:ours_interrupt}
     \end{subfigure}
     \begin{subfigure}[b]{.233\textwidth}
         \centering
         \includegraphics[width=\textwidth]{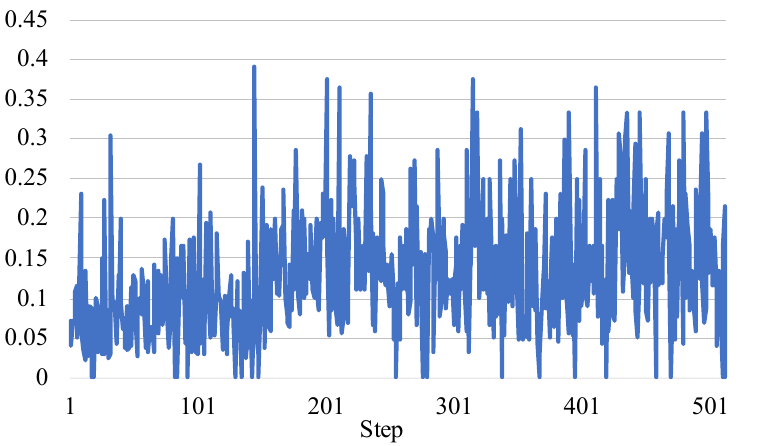}
         \caption{Response Score (ASPIRin)}
         \label{fig:ours_response}
     \end{subfigure}
        
    \caption{Comparison of training reward dynamics between standard GRPO and ASPIRin.}
    \label{fig:six_total_figures}
\end{figure}

\subsection{Analysis of Reward Dynamics}

Standard GRPO and ASPIRin both display an upward trend in total reward throughout training, yet their Interruption Score dynamics differ dramatically. As shown in Figures \ref{fig:grpo_interrupt} and \ref{fig:ours_interrupt}, standard GRPO exhibits severe instability, featuring rapid oscillations and a consistent downward trend that signals clear degradation, while ASPIRin preserves stable Interruption Score values throughout the training process (Figure \ref{fig:ours_interrupt}) without observable degradation. This problematic behavior in standard GRPO severely undermines the reliability of using loss or total reward convergence as a criterion for terminating training.

The Interruption Score trajectories explain the main behavioral gaps. Standard GRPO overprioritizes response scores, ignoring interruption costs and causing severe TOR degradation in pauses and backchannels. ASPIRin, which advances more conservatively, balances both constraints and delivers stable TOR gains.
This advantage stems directly from Action Space Projection: mapping to a binary “speak or not” decision concentrates learning on timing alone. The model thus discovers that silence can be rewarding, enabling effective full-duplex optimization.

\begin{table}[t]
    \centering
    \caption{\textbf{Qualitative examples from the "User Interruption" task.} While the Standard SFT baseline hallucinates irrelevant vocabulary and Standard GRPO suffers from severe repetitive loops, ASPIRin successfully maintains semantic coherence and contextually appropriate responses, achieving parity with the base Moshi model.}
    \label{tab:qualitative}
    \scriptsize
    \setlength{\tabcolsep}{3pt} 
    \begin{tabularx}{\linewidth}{l X} 
        \toprule
        & \textbf{Example} \\ 
        \midrule
        \textbf{Input Question} & By the way, how about we discuss exercise routines instead? \\ 
        \midrule 
        \textbf{Moshi} & Sure, I'm good with that. What's your routine look like? I've been trying to incorporate more cardio and strength framing into my routine. I've been doing a mix of running. \textbf{(Rating: 4)} \\ 
        \addlinespace
        \textbf{Standard SFT} & Joyful Veal Eh is beneficial But I work in diverse inventions, so this actual definite divention of grapefish individual ... \textbf{(Rating: 0)} \\ 
        \addlinespace
        \textbf{Standard GRPO} & Hey, so I was just thinking, I have this friend, you know, who has a dog and you know, he says that his dog is uh you know really good you know at tracking and you know I was just wondering you know what you know what kind of you know ... \textbf{(Rating: 0)} \\ 
        \addlinespace
        \textbf{ASPIRin} (Ours) & Sure, that sounds good. Let's start with the basic question. What are some of the most common types of exercises? Okay, so there are a lot of different types of exercises, but some of the most common ones are ... \textbf{(Rating: 5)} \\ 
        \bottomrule
    \end{tabularx}
\end{table}

\begin{table}[t]
    \centering
    \caption{\textbf{Evaluation of degenerative repetition using seq-rep-n and Self-BLEU.} By isolating timing optimization from semantic token selection, ASPIRin effectively mitigates the degenerative repetition loops observed in standard GRPO, reducing 2-gram and 3-gram repetition by over 50\%.}
    \setlength{\tabcolsep}{1.13mm}{
    \scriptsize
    \begin{tabular}{lcccccccc}
        \toprule
        \multirow{2}{*}[-0.5ex]{\textbf{Metric}} & \multicolumn{3}{c}{\textbf{seq-rep-n}} & \multirow{2}{*}[-0.5ex]{\textbf{Self-BLEU} ($\downarrow$)} \\
    \cmidrule(lr){2-4} 
    & 1-gram ($\downarrow$) & 2-gram ($\downarrow$) & 3-gram ($\downarrow$) & \\
        \midrule
        
        Standard GRPO & 0.303 & 0.117 & 0.072 & 0.369 \\ 
        ASPIRin (Ours) & \textbf{0.202} & \textbf{0.054} & \textbf{0.029} & \textbf{0.343} \\
        \bottomrule
    \end{tabular}}
    \vspace{-5mm}
    \label{tab:repetition}
\end{table}

\subsection{Analysis of Semantic Quality and Repetition}

To investigate the discrepancies in GPT-4o semantic ratings, we qualitatively analyze examples from the "User Interruption" task (Table \ref{tab:qualitative}). Both the base Moshi model and ASPIRin produce contextually appropriate responses and consistently receive ratings between 4 and 5. In stark contrast, standard GRPO fails completely. Its outputs are not only meaningless but also heavily affected by repetitive patterns, which is a well-documented symptom of generative degradation \cite{Welleck2020Neural,zhu2018texygen}.

To quantify this degradation, we measure the severity of repetition using the portion of duplicate n-grams (seq-rep-n) and Self-BLEU for assessing intra-sequence repetition and inter-sample diversity, respectively. The corresponding quantitative results are presented in Table \ref{tab:repetition}. The empirical metrics perfectly align with our qualitative observations: standard GRPO exhibits severe generative collapse, yielding high repetition scores across all metrics. Crucially, ASPIRin effectively mitigates this issue, generating significantly more diverse content. Specifically, ASPIRin cuts 2-gram and 3-gram overlap by more than half compared to standard GRPO, and reduces the overall Self-BLEU score from 0.369 to 0.343. These findings confirm that isolating timing optimization from semantic token selection prevents the degenerative repetition characteristic of standard raw-token RL.

\section{Conclusion}

We introduced ASPIRin, an interactivity-optimized reinforcement learning framework resolving the tension between temporal dynamics and semantic coherence in full-duplex SLMs. While standard GRPO burdens fine-grained token policies and suffers from aggressive, repetitive generation, ASPIRin utilizes Action Space Projection to map vocabulary into a binary active/inactive state. Optimizing this coarse-grained policy with rule-based rewards successfully balances prompt responsiveness with low interruption risk. Evaluations confirm ASPIRin outperforms standard GRPO across diverse conversational scenarios without sacrificing base linguistic quality.

Future work will investigate more expressive action spaces beyond the current binary “speak or not” decision. For instance, we can distinguish backchannel utterances (e.g., “uh-huh”) as a dedicated class separate from full responses or interruptions. Such multi-class or hierarchical designs could enable finer-grained control over timing and content, facilitating the development of more natural and interactive full-duplex systems.

\section{Generative AI Use Disclosure}
During the preparation of this work, the authors used Generative AI tools exclusively for editing and polishing the manuscript to improve overall readability. Generative AI was not used to produce any significant portion of the manuscript's original content, ideas, or research findings. All co-authors consent to this submission, take full responsibility and accountability for the final content of this paper, and confirm that no Generative AI tool is listed as a co-author.

\section{Acknowledgements}
We thank the ASUS Open Cloud Infrastructure Software Center for providing the essential resources that supported this work. We are also grateful to Steve Chung-Cheng Chen, Tsung-Ying Yang, Jen-Hao Cheng, and Dau-Cheng Lyu for their insightful discussions and feedback. Additionally, this research was supported by the National Center for High-Performance Computing (NCHC) of the National Applied Research Laboratories (NARLabs), Taiwan, whose advanced infrastructure and academic resources were instrumental to the completion of this study.

\bibliographystyle{IEEEtran}
\bibliography{mybib}

\end{document}